\documentclass[letterpaper, 10 pt, journal, twoside]{IEEEtran}
\usepackage{graphics} % for pdf, bitmapped graphics files
\usepackage{epsfig} % for postscript graphics files
\usepackage{amsmath} % assumes amsmath package installed
\usepackage{amssymb}  % assumes amsmath package installed
\usepackage{algorithm}
\usepackage{algpseudocode}
\usepackage{multirow}
\usepackage{cite}
\makeatletter
\let\NAT@parse\undefined
\makeatother
\usepackage[
            citecolor=black,
            ]{hyperref}
\usepackage{booktabs}
\usepackage{tabularx} % 
\usepackage{lineno,hyperref}
\usepackage{color,xcolor}
\usepackage[inline]{enumitem}
\usepackage{tabularx}

\usepackage{amsmath}
\renewcommand{\thetable}{\Roman{table}}  

\newcommand{\name}{\textsc{CausalNav}}
\usepackage{placeins}  

\begin{document}

\title{\name{}: A Long-term Embodied Navigation System for Autonomous Mobile Robots in Dynamic Outdoor Scenarios}

\author{
Hongbo Duan$^{1}$, Shangyi Luo$^{1}$, Zhiyuan Deng$^{1}$, Yanbo Chen$^{1}$,   Yuanhao Chiang$^{1}$, \\
Yi Liu$^{1}$, Fangming Liu$^{2}$, Xueqian Wang$^{1}$
\thanks{Manuscript received: June, 11, 2025; Revised November, 6, 2025; Accepted January, 2, 2026.}
\thanks{This paper was recommended for publication by Editor Ashis Banerjee upon evaluation of the Associate Editor and Reviewers' comments.
This work was supported by the National Natural Science Foundation of China under Grant Nos. 62293545 and U21B6002, in part by the Major Key Project of PCL under Grant PCL2024A06 and PCL2025A10, and in part by the Shenzhen Science and Technology Program under Grant RCJC20231211085918010.
\textit{(Corresponding author: Xueqian Wang.)}} 
\thanks{$^{1}$ Center for Artificial Intelligence and Robotics, Shenzhen International\\ Graduate School, Tsinghua University, Shenzhen 518055, China. {\tt\footnotesize dhb24@\\mails.tsinghua.edu.cn;
wang.xq@sz.tsinghua.edu.cn}}

\thanks{$^{2}$ Peng Cheng Laboratory, 518108, China. 
{\tt\footnotesize fangminghk@gmail.com}}

\thanks{Digital Object Identifier (DOI): see top of this page.}
}

\markboth{IEEE Robotics and Automation Letters. Preprint Version. Accepted January, 2026}
{Duan \MakeLowercase{\textit{et al.}}: CausalNav: A Long-term Embodied Navigation System for Autonomous Mobile Robots in Dynamic Outdoor Scenarios} 

\maketitle

\begin{abstract}

Autonomous language-guided navigation in large-scale outdoor environments remains a key challenge in mobile robotics, due to difficulties in semantic reasoning, dynamic conditions, and long-term stability. We propose CausalNav, the first scene graph-based semantic navigation framework tailored for dynamic outdoor environments. We construct a multi-level semantic scene graph using LLMs, referred to as the \textit{Embodied Graph}, that hierarchically integrates coarse-grained map data with fine-grained object entities. The constructed graph serves as a retrievable knowledge base for Retrieval-Augmented Generation (RAG), enabling semantic navigation and long-range planning under open-vocabulary queries. By fusing real-time perception with offline map data, the \textit{Embodied Graph} supports robust navigation across varying spatial granularities in dynamic outdoor environments. Dynamic objects are explicitly handled in both the scene graph construction and hierarchical planning modules. The \textit{Embodied Graph} is continuously updated within a temporal window to reflect environmental changes and support real-time semantic navigation. Extensive experiments in both simulation and real-world settings demonstrate superior robustness and efficiency. 
\end{abstract}

\IEEEpeerreviewmaketitle

\section{Introduction}

\IEEEPARstart{R}{ecent} advances in mobile robotics have shifted the focus beyond traditional control, perception, and navigation. Lifelong navigation demands not only accuracy, but also semantic understanding, incremental learning in dynamic environments, long-term robustness, and decision-making capabilities~\cite{1}. This calls for deeper integration between robotics and AI to enable scalable and adaptive autonomy.

Autonomous navigation in large-scale outdoor environments remains challenging due to their dynamic and unpredictable nature~\cite{2,yin2023bioslam}. Robots must perform both local and global planning while reasoning about semantics and high-level task intent. However, most visual-language navigation (VLN) benchmarks are confined to static indoor settings with step-by-step instructions, diverging from real-world scenarios where humans provide abstract goals and expect robots to understand semantics, infer spatial relations, and navigate from arbitrary start points~\cite{3}.

\begin{figure}[t]
\centering
\includegraphics[scale=0.75]{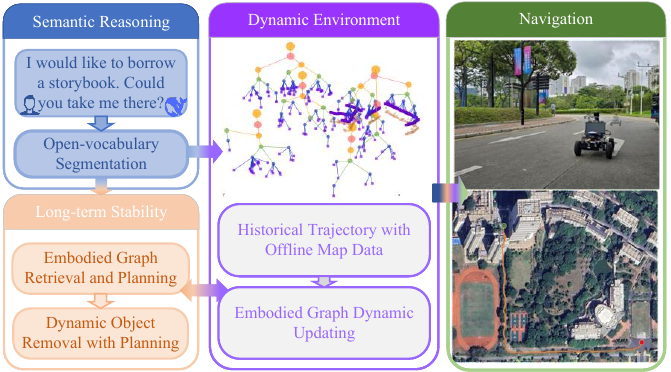}
\caption{The overall workflow of CausalNav. CausalNav introduces a novel embodied navigation framework that integrates open-vocabulary semantic reasoning, dynamic environment adaptation, and Embodied Graph-based planning. By dynamically updating the navigation graph with both historical and real-time data, the system enables robust, long-horizon, and language-directed navigation in complex outdoor environments. }
\label{figure:1}
\end{figure}

Traditional outdoor navigation methods are primarily point-to-point and rely heavily on high-precision map construction\cite{4}. However, most existing studies in this area offer limited or shallow semantic understanding and reasoning in open-world environments\cite{shah2022rapid,Shah-RSS-22}, making sustained human–robot interaction difficult. Although learning-based navigation methods have advanced rapidly, their robustness in long-range, real-world scenarios remains insufficiently validated\cite{sridhar2024nomad,liu2024citywalker}. They also struggle with dynamic environments and often require large-scale training data. Most visual-language navigation (VLN) research is still conducted in indoor settings\cite{5,rana2023sayplan,huang2023visual,kassab2024language}, with limited ability to handle environmental changes. In contrast, outdoor environments are large-scale, highly dynamic, and often degraded, posing greater challenges for embodied navigation and demanding higher levels of algorithmic robustness.

To address the above challenges, we propose CausalNav, an open-world semantic navigation system enhanced with offline map data. CausalNav leverages LLM-constructed scene graphs enhanced by RAG retrieval, coupled with hierarchical global-local planning, to achieve open-vocabulary long-range navigation in dynamic outdoor environments.
Our framework extends recent efforts in applying RAG to robotics~\cite{xie2024embodied,booker2024embodiedrag,wang2025navrag}, introducing key improvements. By leveraging urban map data and real-time perception, we construct a multi-level Embodied Graph that integrates coarse-grained buildings and fine-grained object entities in urban campus environments. The graph is updated dynamically and supports multi-scale memory representations of navigation targets, while also incorporating dynamic objects within a temporal window. In addition, hierarchical summarization using LLMs improves the spatial abstraction capabilities of the Embodied Graph by organizing scene topology into semantically structured layers.

As illustrated in Fig. \ref{figure:1}, our navigation framework integrates open-vocabulary semantic reasoning, dynamic environment adaptation, and Embodied Graph-based planning, which together enable language-directed, long-horizon, and robust navigation across dynamic outdoor scenes. Notably, we demonstrate that deploying LLMs at the edge enables robust performance in urban campus environments without relying on commercial APIs (e.g., GPT-4o). Open-source models running locally on autonomous platforms offer a practical and effective alternative. In summary, our key contributions are as follows:

\begin{enumerate}
    \item We construct a multi-level \textit{Embodied Graph} with LLMs, integrating hierarchical semantics from coarse-grained map data to fine-grained object entities, serving as a RAG-retrievable knowledge base for semantic navigation. 
    
    \item We address dynamic object handling by continuously updating \textit{Embodied Graph} and voxel maps at both the scene graph and planning levels. The graph is dynamically updated within a temporal window to support robust retrieval across different spatial scales.
    
    \item We propose \textit{CausalNav}, the first scene graph-based semantic navigation framework tailored for dynamic outdoor environments. We validate the system in both simulation and real-world campus scenarios. Compared with state-of-the-art methods, CausalNav demonstrates superior performance in terms of navigation success rate, trajectory efficiency, and robustness.
\end{enumerate}

\section{Related Works}

\subsection{Traditional Outdoor Navigation Methods}

Achieving reliable and efficient navigation for ground robots in complex outdoor environments remains a long-standing challenge~\cite{2}.  
Traditional pipelines decompose navigation into independent modules—perception~\cite{3}, localization~\cite{cadena2016past}, path planning, and control—offering interpretability and robustness.  
{However, their reliance on high-definition (HD) maps and manual calibration limits scalability in large and dynamic environments.}  
{While suitable for structured scenarios without large-scale data or end-to-end training~\cite{4}, these methods lack generalization to open-ended tasks and cannot handle natural language interaction.}  
{Their dependence on static maps and handcrafted rules further constrains adaptability to real-world dynamic scenes.}

\subsection{Learning-based Navigation Methods}

Recent advances in learning-based navigation, particularly reinforcement learning (RL), have enabled direct policy learning from visual inputs\cite{hao2023exploration}. Methods such as RECON~\cite{shah2022rapid} and ViKiNG~\cite{Shah-RSS-22} incorporate latent goal models, topological memory, and traversability estimation to support exploration and local planning. Meanwhile, diffusion models have emerged for local trajectory generation, with approaches like NoMaD~\cite{sridhar2024nomad} and ViNT~\cite{shah2023vint} applying diffusion-based action or goal sampling for efficient planning and obstacle avoidance~\cite{carvalho2023motion}. Despite their innovation, these methods are typically constrained to short-range navigation and lack explicit global path reasoning. Moreover, they often require large scale environment-specific training data and show limited generalization due to the sim-to-real gap. Their robustness in dynamic or long-horizon real-world scenarios remains insufficiently validated.

\subsection{Scene Graph-based Navigation Methods}

Scene graph-based visual-language navigation has rapidly progressed, especially in indoor settings~\cite{5,rana2023sayplan,huang2023visual,kassab2024language}.  
{Scene graphs provide high-level spatial abstractions that decouple perception from sensors and support hierarchical reasoning.}  
Hydra~\cite{5} constructs hierarchical 3D scene graphs from metric to semantic levels, while VLMaps~\cite{huang2023visual} fuses pretrained vision-language features with 3D maps.  
{LExis~\cite{kassab2024language} aligns visual and language representations for semantic SLAM and topological planning.}  
{However, most of these methods remain constrained to indoor environments, where outdoor scale and dynamics pose major challenges.}

{With the rise of foundation models, recent works integrate LLMs for language-conditioned planning.}  
SayPlan~\cite{rana2023sayplan} combines LLMs with 3D scene graphs for scalable task reasoning, and LM-Nav~\cite{shah2023lm} enables instruction-following for novel object goals.  
{Yet, these systems often treat LLMs or VLMs as isolated evaluators rather than components of an integrated mapping framework.}  
{Moreover, hallucination and forgetting limit their reliability in dynamic, real-time outdoor scenarios.}

\begin{figure*}[t]
  \centering
  {\includegraphics[scale=0.72]{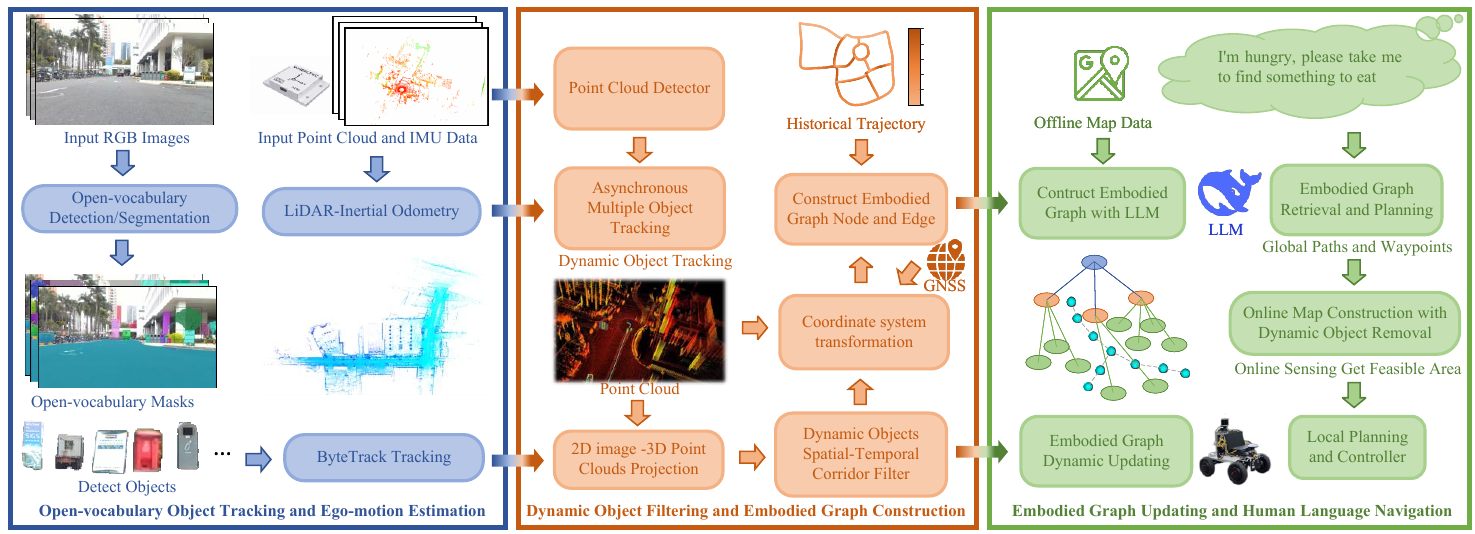}}
  \caption{ The CausalNav framework comprises three sequential modules:
(1) Open-vocabulary Object Tracking and Ego-motion Estimation(in Sec. \ref{Open-vocabulary Object Tracking and Ego-motion Estimation}): Integrates RGB, LiDAR, and IMU inputs for open-vocabulary object detection and tracking, along with ego-motion estimation via 2D–3D spatial-temporal alignment.
(2) Dynamic Object Filtering and Embodied Graph Construction(in Sec. \ref{retrival}): Filters transient dynamic objects through asynchronous multi-object tracking and constructs a temporally stable embodied scene graph.
(3) Embodied Graph Updating and Human Language Navigation(in Sec. \ref{Embodied Graph Updating and Language-guided Planning}): Builds and updates a semantic graph using LLMs to interpret language commands and perform hierarchical planning, with real-time dynamic object removal for robust navigation. }
  \label{figure2}
  \vspace{-6pt}
\end{figure*}

\section{Methodology}
\subsection{System Overview}

As shown in Fig. \ref{figure2}, CausalNav adopts a hierarchical architecture comprising perception, graph construction, and planning components, tailored for long-horizon navigation in dynamic outdoor environments. Inspired by OpenGraph\cite{deng2024opengraph}, the system constructs a multi-level Embodied Graph \(G\) composed of object nodes \(\nu_i^{obj}\), ego-vehicle nodes \(\nu_i^{l}\), building nodes \(\nu_i^{build}\), and hierarchical clustering nodes \(\nu_i^{cluster}\). This graph integrates coarse-grained map information with fine-grained semantic entities. Its detailed structure is depicted in Fig. \ref{figure:sim_environ}. In Sec. \ref{Open-vocabulary Object Tracking and Ego-motion Estimation} semantic alignment is achieved by leveraging an open-vocabulary perception model~\cite{cheng2024yolo} and LiDAR-based localization~\cite{xu2022fast}, enabling the extraction of object and ego-vehicle node information. 

In Sec. \ref{retrival}, the system distinguishes dynamic, static, and quasi-static objects through multi-object tracking. To mitigate the impact of environmental dynamics on the construction of the Embodied Graph, a spatio-temporal filtering mechanism is applied before updating the semantic graph. By introducing building nodes and performing hierarchical clustering with the assistance of large language models (LLMs), the constructed Embodied Graph supports retrieval-augmented generation, enabling the system to respond effectively to human language instructions. 

In Sec. \ref{Embodied Graph Updating and Language-guided Planning}, the system continuously integrates offline maps with live perception data to update the graph in real time. Global navigation relies on either pre-stored routes or dynamically generated scene graphs, whereas local path planning is adaptively performed using real-time LiDAR mapping and on-the-fly removal of dynamic obstacles. This design enables flexible, language-driven navigation with long-term stability and semantic awareness.

\subsection{Open-vocabulary Object Tracking and Ego-motion Estimation} 
\label{Open-vocabulary Object Tracking and Ego-motion Estimation}
\subsubsection{Open-vocabulary Object Tracking} To obtain rich information about the description, shape, size, and position of object nodes, we denote the object state as \( {}^{w} \mathbf{T}_{\text{obj}} = \left\{ {}^{w} \mathbf{R}_{\text{obj}}, {}^{w} \mathbf{p}_{\text{obj}} \right\} \), where the object's pose \( { }^{w} \mathbf{T}_{obj} \)  represented in the \( SE(3) \) space.
In outdoor environments, we use YOLO-World~\cite{cheng2024yolo}, a lightweight open-vocabulary detector, to extract 2D bounding boxes and segmentation masks from each RGB frame \( I_t \). These detections are then temporally associated by the multi-object tracking function \( \mathcal{C}(\cdot) \), implemented by ByteTrack~\cite{zhang2022bytetrack}, yielding stable tracking results:
\begin{equation}
    \mathcal{S}_t = \mathcal{C}(\text{YOLO-World}(I_t)).
\end{equation}
% S_t = f(\text{YOLO-World}(I_t))

Here, \( \mathcal{S}_t \) denotes the set of detected objects in the current frame, defined as
\begin{equation}
    \mathcal{S}_t = \left\{ S_i = \left(  c_i, \text{2DBBox}_i,\mathcal{B}_i\right) \mid i = 1, 2, \ldots, n \right\},
\end{equation}
where \( \mathcal{B}_i \) is the segmentation mask, \( \mathrm{2DBBox}_i \) the 2D bounding box, and \( c_i \) the description for the \(i\)-th tracked object.

Due to the limited performance of depth cameras in real outdoor environments, we fuse camera and LiDAR data for object localization. The raw LiDAR point cloud is denoted as \( \mathcal{P} = \left\{ \mathbf{P}_i = (x_i, y_i, z_i) \right\}_{i=1}^N \). For the current time step \( t \), each LiDAR point \( \mathbf{P}_i \in \mathcal{P}_t \) is projected onto the image plane of frame \( I_t \) via the calibrated camera projection model, yielding image-space coordinates 
\({}^{c} \mathbf{p}_i\), computed as:
\(
{}^{c} \mathbf{p}_i
 = \mathbf{K} \cdot \mathbf{H} \cdot \mathbf{P}_i,
\)
where \( \mathbf{K} \) is the camera intrinsic matrix and \( \mathbf{H} \) is the extrinsic transformation matrix obtained through joint LiDAR-camera calibration~\cite{tsai2021optimising}.

We then define the object-specific 3D point cloud in the camera frame as:
\begin{equation}
{}^{l}\mathcal{P}_{\text{obj}} = \left\{ \mathbf{P}_i \in \mathcal{P}_t \,\middle|\, {}^{c} \mathbf{p}_i \in \mathcal{B}_i \right\}.
\end{equation}

Based on the filtered set \( {}^{l}\mathcal{P}_{\text{obj}} \), a minimum-volume 3D bounding box is constructed. The centroid of this bounding box is taken as the estimated 3D position of the object, denoted as \( {}^{l} \mathbf{p}_{\text{obj}} \).

We obtain the current ego-pose \( {}^{w} \mathbf{T}_{l} \) from a LiDAR-inertial odometry system\cite{xu2022fast}. The relative pose of the detected object in the LiDAR frame is denoted as \( {}^{l} \mathbf{T}_{\text{obj}} \). The corresponding world-frame object pose is calculated by \(
{}^{w} \mathbf{T}_{\text{obj}} = {}^{w} \mathbf{T}_{l} \cdot {}^{l} \mathbf{T}_{\text{obj}}\). For incremental topological graph updates, each tracked object contributes a node defined as:

\begin{equation}
\nu_i^{obj} = \left\{ c_i, \text{3DBBox}_i, {}^{w} \mathbf{p}_{\text{obj}} \right\},
\end{equation}
which is inserted or updated into the Embodied Graph \( G \).

For each tracked object \( \nu_i^{obj} \), if it is not in the graph \( G \), a new node is created; otherwise, its position is updated:
\begin{equation}
G \leftarrow 
\begin{cases} 
G \cup \{\nu_i^{obj}\}, & \text{if } \nu_i^{obj} \notin G \\
G \setminus \{{}^{old}\nu_i^{obj}\} \cup \{\nu_i^{obj}\}, & \text{if } {}^{old}\nu_i^{obj} \in G. 
\end{cases}
\label{eq:update_rule}
\end{equation}
\subsubsection{Ego-motion Estimation}
\label{Ego-motion Estimation}
For the ego-vehicle, its state is represented as 
\({}^{w} \mathcal{E}_{l} = \left[ {}^{w} \mathbf{R}_{l},\ {}^{w} \mathbf{p}_{l},\ {}^{w} \boldsymbol{\xi}_{l} \right]\), 
where \({}^{w} \mathbf{R}_{l}\) denotes orientation, \({}^{w} \mathbf{p}_{l}\) the position, and 
\({}^{w} \boldsymbol{\xi}_{l}\) the velocity in the world frame. 
When the movement step exceeds a threshold \( d \), the position and velocity are stored as an 
ego-vehicle node 
\(\nu_i^{l} = \left\{ {}^{w} \mathbf{p}_{l},\ {}^{w} \boldsymbol{\xi}_{l} \right\}\), 
and a new edge 
\(e_i^{l} = \left( \nu_i^{l},\ \nu_{i-1}^{l} \right)\) 
is added to \(G\).
As shown in Fig.~\ref{figure:sim_environ}, the ego-vehicle nodes \(\nu_i^{l}\) and edges \(e_i^{l}\), representing the historical trajectory, 
are stored in the Embodied Graph and later utilized for the global planning task described in Sec.~\ref{Human Language Navigation}.

\subsection{Dynamic Object Filtering and Embodied Graph Construction}
\label{retrival}
\subsubsection{Dynamic Objects Spatial-Temporal Corridor Filter}
To robustly handle dynamic objects in outdoor environments, we employ a multi-object tracking pipeline that integrates BEV-based detection with motion estimation. Specifically, CenterPoint~\cite{yin2021center} performs real-time point cloud detection, accelerated via TensorRT, while LIOsegmot~\cite{lin2023asynchronous} estimates object velocities by fusing detections with LiDAR-inertial odometry. This enables reliable classification of objects as dynamic, static, or quasi-static.

\begin{figure}[t]
  \centering
  \includegraphics[scale=0.5]{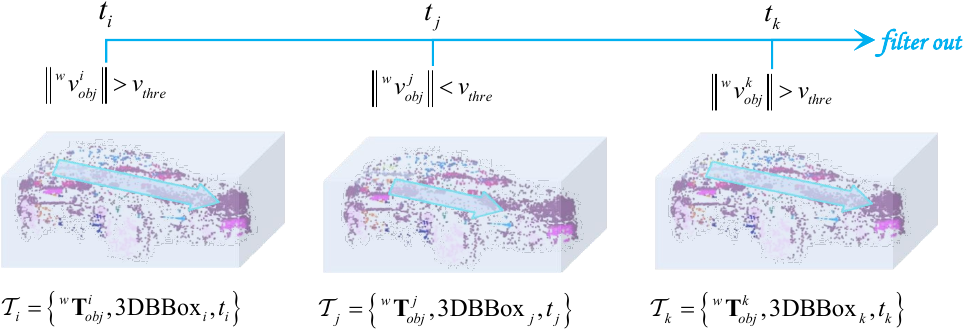}
  \caption{Illustration of three observed trajectory points and their corresponding 3D bounding boxes within the spatial-temporal corridor. The point cloud depicts the same vehicle captured at different timestamps.}
  \label{figure:filter}
  % \vspace{-6pt}
\end{figure}

To mitigate false positives in the Embodied Graph caused by traditional velocity-based filters~\cite{TRLO1}, we encode each object’s trajectory as a spatial-temporal corridor using historical bounding boxes, as illustrated in Fig.~\ref{figure:filter}. This representation enhances the filtering of dynamic entities and preserves the structural integrity of the graph.

\begin{equation}
\label{patiotemporal_corridor}
\mathcal{T} = \left\{ {}^{w} \mathbf{T}_{\text{obj}}^{i}, \text{3DBBox}_i, t_i \right\}_{i=1}^{n}.
\end{equation}

When an object exceeds a displacement threshold of \( k \) steps, its spatial-temporal corridor \(\mathcal{T}\) is excluded, and the corresponding dynamic nodes \( {}^{d}\nu_i^{\text{obj}} \) are removed from graph construction:
\begin{equation}
    G \leftarrow G \setminus \{ \mathcal{T} \mid \mathcal{T} \in D \}.
\end{equation}

This corridor-based filtering effectively removes transient or mobile objects from the Embodied Graph, reducing motion-induced errors. It is particularly effective for entities with intermittent motion patterns, such as vehicles near intersections.

\subsubsection{Hierarchical Construction of the Embodied Graph}

{Beyond the ego-vehicle nodes described in Sec.~\ref{Ego-motion Estimation}, 
we construct a hierarchical spatial–semantic representation of the static environment 
after filtering transient dynamic entities via the spatial–temporal corridor.}
The resulting Embodied Graph integrates geometric and semantic information, 
{serving as the foundation for downstream reasoning and navigation.}

The static topological graph includes two node types:  
(1) building nodes \( \nu_i^{\text{build}} = \{ c_i^{\text{build}}, {}^{w}\mathbf{p}_{\text{build}} \} \), 
extracted from offline maps and assigned the highest level \(L\); and  
(2) object nodes \( \nu_i^{\text{obj}} \), representing fine-grained elements 
(e.g., fire hydrants, bus stops) at level \(L{-}1\).  
{Together, they form a multi-level spatial–semantic abstraction of the environment.}

Following~\cite{booker2024embodiedrag}, hierarchical clustering is performed 
with a spatial–semantic similarity measure:
\begin{equation}
\kappa_{ij} = (1-\alpha)\kappa_{ij}^{\text{spatial}} + \alpha\kappa_{ij}^{\text{semantic}},
\end{equation}
where \( \kappa_{ij}^{\text{spatial}} = \exp(-d_{\text{haversine}}(i,j)/\theta) \), 
and \( \kappa_{ij}^{\text{semantic}} \) is the cosine similarity between embeddings \( \mathbf{e}_i \) and \( \mathbf{e}_j \).  
{Embedding-based similarity enhances robustness to LLM label variations 
(e.g., “trash can” vs. “garbage bin”).}

{Level-\(L{-}1\) object nodes are grouped bottom-up to form level-\(L\) clusters 
\(\nu_i^{\text{cluster}} = \{\text{LLM}(r), \text{centroid}(r)\}\).}
Each cluster \(r \in \mathcal{R}\) contains several nodes, 
with its semantics summarized by an LLM and position given by the mean of its members.  
{Clustering proceeds recursively, jointly considering cluster and building nodes, until convergence.}

\subsubsection{Semantic Retrieval over the Embodied Graph}
\label{Semantic Retrieval over the Embodied Graph}

Given a query \(q\), semantic retrieval is performed hierarchically using LLM-based selection.  
At level \(l\), the selection probability of node \(n_l \in \mathcal{L}_l\) is:
\begin{equation}
\pi(n_l \mid q) =
\frac{\exp[\gamma \cdot \mathrm{LLM}(q, C(n_l))]}
{\sum_{n'\in\mathcal{L}_l}\exp[\gamma \cdot \mathrm{LLM}(q, C(n'))]},
\end{equation}
where \(C(n_l)\) denotes the node description and \(\gamma>0\) controls sharpness.

{The hierarchical path score is:}
\begin{equation}
\Lambda(\zeta)=\prod_{l=1}^{D}\left[\pi(n_l\mid q)\cdot\phi(n_l,n_{l-1})\right],
\end{equation}
where \(\phi(n_l,n_{l-1})=\mathbf{1}_{\{n_{l-1}\in\mathrm{Children}(n_l)\}}\)
ensures valid parent–child links.

If the agent’s location \(\mathbb{L}\) is known, candidates are re-ranked by a hybrid score:
\begin{equation}
\eta(n)=\beta\kappa^{\mathrm{spatial}}(n,\mathbb{L})+(1-\beta)\Lambda(\zeta),
\end{equation}
where \(\kappa^{\mathrm{spatial}}(n,\mathbb{L})=\exp[-d_{\mathrm{haversine}}(n,\mathbb{L})/\theta]\)
measures spatial proximity, and \(\beta\!\in\![0,1]\) balances spatial and semantic relevance.  

{Evaluating \(\eta(n)\) across candidate paths allows the system to select nodes 
most aligned with the query both semantically and spatially for downstream planning.}

\subsection{Embodied Graph Updating and Human Language Navigation}
\label{Embodied Graph Updating and Language-guided Planning}
% ### Dynamic Object Filtering
\subsubsection{Online Embodied Graph Updating} Essentially, the \textit{Embodied Graph} serves as both a Scene Operator and a Memory Tank, abstracting elements such as time, scenes, objects, and events into structured memory representations. Its core value lies in enabling robots to understand complex environments—that is, to perform a form of abstract reasoning—which facilitates more intelligent and diverse decision-making in navigation tasks. During the online construction and updating of the Embodied Graph, we focus on the following aspects:

\begin{itemize}
    \item Multi-granularity topological representation of scenes;
    \item Spatial relationships among static objects within a single scene;
    \item Modeling and management of dynamic objects over temporal windows;
    \item Implicit encoding of inter-scene relationships.
\end{itemize}

% ### Next Steps

The online update process of the Embodied Graph is formally described in Algorithm~\ref{alg:eg_update}.

\begin{algorithm}[t]
\caption{Online Embodied Graph Updating}
\label{alg:eg_update}
\begin{algorithmic}[1]
\Require $\mathcal{C}$, $k$
\State $G \gets \emptyset,\ t \gets 0$
\While{System is running}
    \State $t \gets t+1$,  $\mathcal{S}_t \gets \mathcal{C}(I_t, \mathcal{P}_t, \text{IMU})$
    \ForAll{$S_i \in \mathcal{S}_t$}
        \State \textbf{Compute} ${}^w\mathbf{T}_{\text{obj}}$, $\mathcal{B}_i$, $\text{3DBBox}_i$
        \State $\nu_i^{obj} \gets \left\{ c_i, \text{3DBBox}_i, {}^{w} \mathbf{p}_{\text{obj}} \right\}$
        \If{$\nu_i^{obj} \notin G$} $G \gets G \cup \{\nu_i^{obj}\}$
        \Else \ \ $G$.\textbf{update}($\nu_i^{obj}$)
        \EndIf
    \EndFor
    \ForAll{${}^{d}\nu_i^{obj} \in  \{\mathcal{T} \mid \mathcal{T} \in D \}$}
        \If{${}^{d}\nu_i^{obj}.\textbf{steps} \geq k  $}
            \State $G \gets G \setminus \{\mathcal{T}\}$
        \EndIf
    \EndFor
    \State \textbf{Update} $\nu_i^{l}$,  $E_{\nu}$ for all $\nu \in G$
    \State $\mathcal{R} \gets \textbf{HCluster}(G)$
    \ForAll{${r} \in \mathcal{R}$}
        \State $E_r \gets \{(\nu_i^{cluster}, \nu_i^{obj}) \mid \nu_i^{obj} \in r\}$
\State $G \gets G \cup \{\nu_i^{cluster}\} \cup E_r$

    \EndFor
\EndWhile
\State \Return $G$
\end{algorithmic}
\end{algorithm}

\subsubsection{Human Language Navigation}
\label{Human Language Navigation}
CausalNav performs navigation in response to human language instructions. Based on the semantic retrieval method described in Sec. \ref{Semantic Retrieval over the Embodied Graph}, the system first infers a global target location from the human language input by leveraging the real-time constructed and updated \textit{Embodied Graph}. If the target is not reachable from the robot’s historical trajectory, a coarse global route is generated using either offline road map data or external APIs such as Google Maps or Amap. If the target is connected via past trajectories, a Dijkstra-based shortest path is computed. The resulting global path is represented as a waypoint sequence \( \mathcal{W} = \{ \mathbf{w}_1, \mathbf{w}_2, \dots, \mathbf{w}_n \} \).

In real-world outdoor environments, robots must operate in highly dynamic scenes. Residual motion trails from dynamic objects often appear as static obstacles on the map, degrading localization and navigation performance. To mitigate this, we adopt RH-Map~\cite{yan2023rh}, a 3D region-wise hash map framework for real-time local mapping and dynamic object removal.

The feasible region for planning that the mobile robot can obtain through RH-Map is denoted as
$\mathcal{F} \subseteq \mathbb{R}^3$. Within \( \mathcal{F} \), an initial trajectory is generated using the informed-RRT*\cite{6942976} algorithm:
\begin{equation}
    \mathcal{Z}_{\text{init}} = \{\mathbf{z}_0, \mathbf{z}_1, \dots, \mathbf{z}_N\}, \quad \mathbf{z}_i = [x_i, y_i, z_i]^T.
\end{equation}

The path \( \mathcal{Z}_{\text{init}} \) is then smoothed via B-spline interpolation. Each interpolated point is assigned an orientation \( \theta_i \) computed from the direction of adjacent waypoints. This yields an oriented reference trajectory:
\begin{equation}
    \mathbf{X}_g = \{ \mathbf{x}_g^0, \mathbf{x}_g^1, \dots, \mathbf{x}_g^M \}, \quad \mathbf{x}_g^i = [x_i, y_i, z_i, \theta_i]^T \in \mathbb{R}^4.
\end{equation}

{Given the current robot state \( \mathbf{x}_t = [x_t, y_t, z_t, \theta_t]^T \) and control input \( \mathbf{u}_t = [v_t, \omega_t]^T \), we formulate a Nonlinear Model Predictive Control with Control Barrier Function\cite{zeng2021enhancing} (NMPC-CBF) problem as a constrained optimization:}
\begin{subequations}\label{eq:nmpc_cbf}
\begin{align}
\min_{\{\mathbf{x}_k, \mathbf{u}_k\}} \sum_{k=0}^{N-1} (& \left\| \mathbf{x}_k - \mathbf{x}_g^k \right\|^2_{Q}
 +  \left\| \mathbf{u}_k \right\|^2_{R} )\label{eq:nmpc_cbf_obj} \\
\text{s.t.} \quad & \mathbf{x}_{k+1} = f(\mathbf{x}_k, \mathbf{u}_k)  \label{eq:nmpc_cbf_dyn}\\
& \mathbf{x}_0 = \mathbf{x}_{\text{init}}  \label{eq:nmpc_cbf_init}\\
& \mathbf{x}_k \in \mathcal{X}, \quad \mathbf{u}_k \in \mathcal{U}, \label{eq:nmpc_cbf_constraints}\\
& { \Delta h_{\text{ob}}^i(\mathbf{x}_k, \mathbf{u}_k) + \lambda_k h_{\text{ob}}^i(\mathbf{x}_k) \geq 0. }
\end{align}
\end{subequations}

{Here, \(\mathcal{X} \subseteq \mathbb{R}^n\) and \(\mathcal{U} \subseteq \mathbb{R}^m\) denote the feasible sets of states and control inputs, respectively, incorporating physical and safety constraints. 
The last constraint incorporates a control barrier function that guarantees the forward invariance of the safe set defined by dynamic obstacles such as pedestrians or vehicles. 
Each barrier function \(h_{\text{ob}}^i(\mathbf{x}) > 0\) defines the safety margin to obstacle~\(i\), evolving with its predicted trajectory, 
typically formulated as 
\(h_i(\mathbf{x}) = (x - x_i^p)^2 + (y - y_i^p)^2 - d_{\text{safe}}^2,\)
where \([x_i^p, y_i^p]^T\) denotes the predicted position of obstacle~\(i\), 
and \(d_{\text{safe}}\) represents the safety radius.}

\section{Experiments and Results}

\subsection{Experiments Setup}

\begin{figure}[!t]
  \centering
  {\includegraphics[scale=0.26]{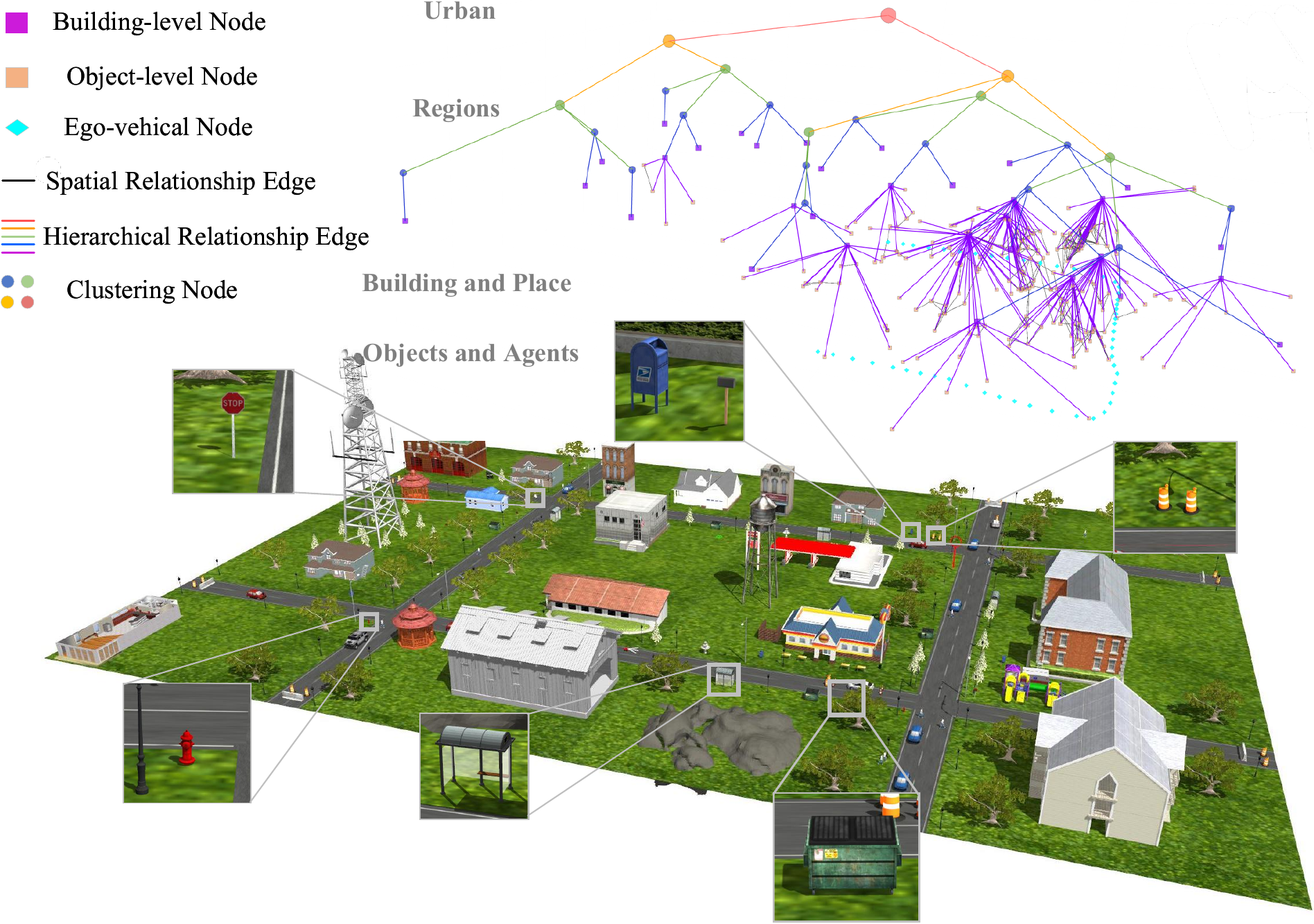}}
  {
  \caption{The simulation environment and the constructed Embodied Graph. The environment includes coarse-grained objects (e.g., buildings) and fine-grained ones (e.g., fire hydrants, mailboxes). The Embodied Graph fuses both levels and updates dynamically as the agent moves.
}
\label{figure:sim_environ}
}
% \vspace{-6pt}
\end{figure}

\textbf{Simulation Environment.}
As shown in Fig.~\ref{figure:sim_environ}, we simulate a Gazebo-based ground robot equipped with a RealSense D435i and 3D LiDAR in an urban environment. The Embodied Graph is dynamically constructed and visualized during navigation. A Gazebo-based state estimator logs task completion and trajectories. Due to the lack of suitable open-source alternatives, we compare our method with four representative learning-based systems: NoMaD, ViNT, GNM, and CityWalker, all of which perform goal-driven motion prediction with obstacle avoidance. For fair comparison, we {collected} image data to build topological maps for NoMaD, ViNT, and GNM, and adopted {our generated} waypoints for CityWalker. Performance is evaluated on 25 randomly sampled tasks (10 trials each) using four metrics: Success Rate (SR), Success weighted by Path Length (SPL), Collision Count (CC), and Trajectory Length (TL). A task succeeds if the robot reaches within 10 meters of the target. The system runs on an Intel i9-14900K CPU and a single RTX 3090 GPU, with key modules operating in real time: {Open-vocabulary object tracking and ego-motion at 30 Hz, spatio-temporal corridor filtering at 20 Hz, local dynamic mapping and planning at 10 Hz, and hierarchical clustering or Embodied Graph updates at 1 Hz.}

\begin{figure}[t]
  \centering
  \includegraphics[scale=0.8]{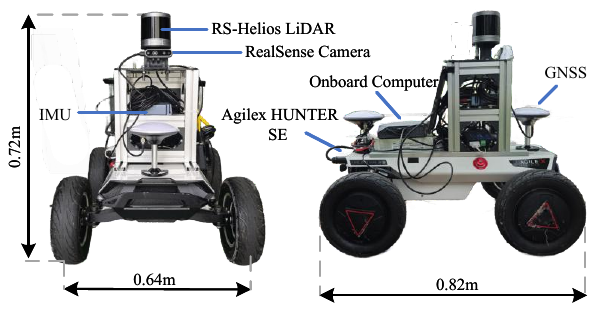}
  \caption{The robot used in real-world navigation experiments.}
  \label{figure:robot}
  % \vspace{-6pt}
\end{figure}

\textbf{Real Environment.} As shown in Fig. \ref{figure:robot}, we deploy our system on a wheeled mobile robot equipped with an Intel Core i9-13900H CPU and an NVIDIA GeForce RTX 4070 GPU. A RealSense D435i camera captures RGB images, while the RSHelios lidar and an RTK GNSS/INS module provide positioning data at 10 Hz with 5 cm accuracy. We utilize FAST-LIO2 \cite{xu2022fast} in conjunction with RTK GNSS/INS for coordinate transformation and precise localization. The system was tested in a large-scale, campus-like outdoor environment to evaluate its effectiveness and robustness in dynamic real-world scenarios.

\subsection{Simulation Environment Experiments}

This experiment {evaluates} our approach in long-range navigation tasks, focusing on success rate, navigation efficiency, and adaptability to dynamic {environments}. {In simulation}, pedestrians and vehicles emulate real-world complexity. According to the spatial scale of the navigation target, tasks are categorized into short-range, medium-range, and long-range scenarios. The results are summarized in Table~\ref{tabel:1}. In terms of SR and SPL, both CityWalker and CausalNav exhibit strong performance. However, analysis of TL in long-range tasks reveals that topological methods such as ViNT, NoMaD, and GNM suffer from unidirectional connectivity, resulting in inefficient path planning and significantly longer trajectories even when the destination is reachable. While CityWalker and CausalNav achieve comparable results on SR, SPL, and TL, they differ significantly in CC. CityWalker, as a learning-based method, is easy to deploy without extensive parameter tuning. However, it shows limited generalization in dynamic environments, often failing to avoid moving obstacles. In contrast, CausalNav enhances dynamic responsiveness by constructing safe zones in real time and performing perception and re-planning within about 100~ms, leading to more robust navigation in dynamic settings. Overall, CausalNav outperforms existing baselines across multiple metrics, demonstrating superior performance in long-range success rate, trajectory efficiency, and dynamic adaptability. 

\textbf{Different LLM on Navigation.}
Large Language Models (LLMs) play a critical role in Embodied Graph construction and semantic reasoning for navigation. 
However, relying on online APIs is impractical in real-world deployments due to limited network access, high latency, and data privacy concerns. 
{This highlights the need for locally deployed LLMs in autonomous navigation.}
We evaluate four recent open-source LLMs with comparable scales, along with the GPT-4o online API, in a simulated environment where the Embodied Graph is built online. 
Navigation performance is summarized in Table~\ref{tab:llm_performance}. 
Although GPT-4o achieves the best performance, {its margin over smaller open-source models is modest,} owing to hierarchical semantic retrieval within the Embodied Graph, which improves query accuracy and mitigates hallucinations. 
Among open-source models, Deepseek-R1-Distill-14B shows the highest success rate and most stable performance.

\begin{table*}[!ht]
\centering
\caption{Comparative analysis of navigation performance across various semantic navigation methods in simulation environment.}
\label{tabel:1}
\renewcommand{\arraystretch}{1.1}
\setlength{\tabcolsep}{1.5pt} % 
% \small
\begin{tabularx}{\textwidth}{@{} 
    >{\raggedright\arraybackslash}p{2cm} % 
    *{12}{>{\centering\arraybackslash}X} 
@{}}
\toprule
\textbf{Method} & 
\multicolumn{4}{c}{\textbf{Small}} & 
\multicolumn{4}{c}{\textbf{Medium}} & 
\multicolumn{4}{c}{\textbf{Large}} \\
\cmidrule(lr){2-5} \cmidrule(lr){6-9} \cmidrule(l){10-13}
& \textbf{SR(\%)$\uparrow$} & \textbf{SPL(\%)$\uparrow$} & \textbf{CC$\downarrow$} & \textbf{TL(m)$\downarrow$} & \textbf{SR(\%)$\uparrow$} & \textbf{SPL(\%)$\uparrow$} & \textbf{CC$\downarrow$} & \textbf{TL(m)$\downarrow$} & \textbf{SR(\%)$\uparrow$} & \textbf{SPL(\%)$\uparrow$} & \textbf{CC$\downarrow$} & \textbf{TL(m)$\downarrow$} \\
\midrule
ViNT \cite{shah2023vint} & \underline{84} & 68.4 & 0.6 & 47.16 & 64 & 47.8 & \underline{0.9} & 92.74 & \underline{48} & 32.2 & \underline{1.6} & 160.03 \\
NoMaD \cite{sridhar2024nomad} & 82 & 70.9 & 0.8 & 45.61 & 43 & 34.2 & 1.3 & \textbf{81.13} & 22 & 14.6 & 2.3 & 173.72 \\
GNM \cite{gnm} & \underline{84} & 72.3 & \underline{0.5} & \underline{42.02} & 26 & 15.2 & 1.5 & 88.32 & 0 & 0 & - & - \\
CityWalker \cite{liu2024citywalker} & \textbf{100} & \underline{82.4} & 1.2 & 43.07 & \underline{85} & \underline{73.6} & 3.4 & 86.32 & \textbf{80} & \textbf{68.3} & 4.5 & \textbf{136.63} \\
\multicolumn{1}{@{}l}{CausalNav} & \textbf{100} & \textbf{88.9} & \textbf{0.2} & \textbf{40.66} & \textbf{92} & \textbf{82.2} & \textbf{0.6} & \underline{83.16} & \textbf{80} & \underline{66.0} & \textbf{1.2} & \underline{141.82} \\
\bottomrule
\end{tabularx}
% 注释部分保持不变
\end{table*}
% 第2个表格
\begin{table}[t]
\centering
\caption{Performance of Different LLMs on Navigation}
\label{tab:llm_performance}
\renewcommand{\arraystretch}{1.1}
\setlength{\tabcolsep}{0pt} % 
\begin{tabular*}{\linewidth}{@{\extracolsep{\fill}}lcccc@{}}
\toprule
\textbf{Method} & \textbf{SR(\%)$\uparrow$} & \textbf{SPL(\%)$\uparrow$} & \textbf{CC$\downarrow$} & \textbf{TL(m)$\downarrow$} \\
\midrule
phi4-14B & 83 & 69.5 & \underline{1.1} & 106.27  \\
Qwen3-14B & 82 & 70.4 & \textbf{1.0} & \textbf{102.45}  \\
Gemma3-12B & 84 & 67.6 & 1.2 & 110.32  \\
DeepSeek-R1-Distill-14B & \underline{85} & \underline{72.1} & \underline{1.1} & 103.63  \\
GPT-4o & \textbf{88} & \textbf{75.3} & \textbf{1.0} & \underline{103.24}  \\
\bottomrule
\end{tabular*}

% \vspace{0.5em}

\end{table}

\begin{table}[t]
\centering
\caption{The Impact of Online Embodied Graph Updating on Navigation}
\label{tab:graph_impact}
\renewcommand{\arraystretch}{1.1}
\begin{tabular}{@{}lcccc@{}}
\toprule
\textbf{Method} & \textbf{SR(\%)$\uparrow$} & \textbf{SPL(\%)$\uparrow$} & \textbf{CC$\downarrow$} & \textbf{TL(m)$\downarrow$} \\
\midrule
CausalNav \textsuperscript{†} & 78 & 54.7 & 1.8 & 120.35 \\
CausalNav \textsuperscript{‡} & \textbf{90} & \textbf{80.1} & \textbf{1.1} & \textbf{98.25} \\
\bottomrule
\end{tabular}

% \vspace{0.5em}
\footnotesize
\textsuperscript{†}Without Embodied Graph updates; \textsuperscript{‡}With Embodied Graph updates. \\
\end{table}

\begin{table}[t]
\centering
\small
{\caption{Mean Per-Cycle Latency Runtime Comparison}
\label{tab:runtime_single}}
{%
\resizebox{0.9\columnwidth}{!}{%
\begin{tabular}{lccccc}
\toprule
\textbf{Method} & NoMaD~\cite{sridhar2024nomad} & ViNT~\cite{shah2023vint} & GNM~\cite{gnm} & CityWalker~\cite{liu2024citywalker} & \textbf{CausalNav} \\
\midrule
Runtime (ms) ↓ & \textbf{95} & 150 & 110 & 180 & \underline{105} \\
\bottomrule
\end{tabular}
}}
% \arrayrulecolor{black}  % 

% \vspace{-4pt}
\end{table}

\textbf{Impact of Dynamic Updates to the Embodied Graph.}
In outdoor environments, localized changes, such as moving pedestrians and vehicles, can introduce temporary occlusions and traversability changes that disrupt the consistency between the Embodied Graph and real-time perception. Although the initial scene graph is constructed through full-environment exploration, it quickly becomes outdated without continuous updates. We evaluated the impact of such updates through a series of continuous navigation tasks with randomly generated destinations, simulating real-world scenarios. As shown in Table~\ref{tab:graph_impact}, the dynamic update strategy yields significant improvements in SR, SPL, CC, and TL metrics compared to the static baseline. These results demonstrate that even without global semantic changes, maintaining up-to-date local information is critical for accurate semantic retrieval and reliable navigation performance.

\textbf{Runtime Efficiency.}
To further evaluate computational efficiency, we compare the average per-cycle latency with representative baselines under identical hardware configurations. As shown in Table~\ref{tab:runtime_single}, CausalNav achieves real-time performance (10~Hz) with only a 11\% overhead over NoMaD, 
demonstrating competitive runtime efficiency while providing richer semantic reasoning and dynamic graph maintenance.

\textbf{Ablation on Semantic Retrieval Parameters.}
\begin{figure}[!htbp]
\centering
{\includegraphics[width=0.98\linewidth]{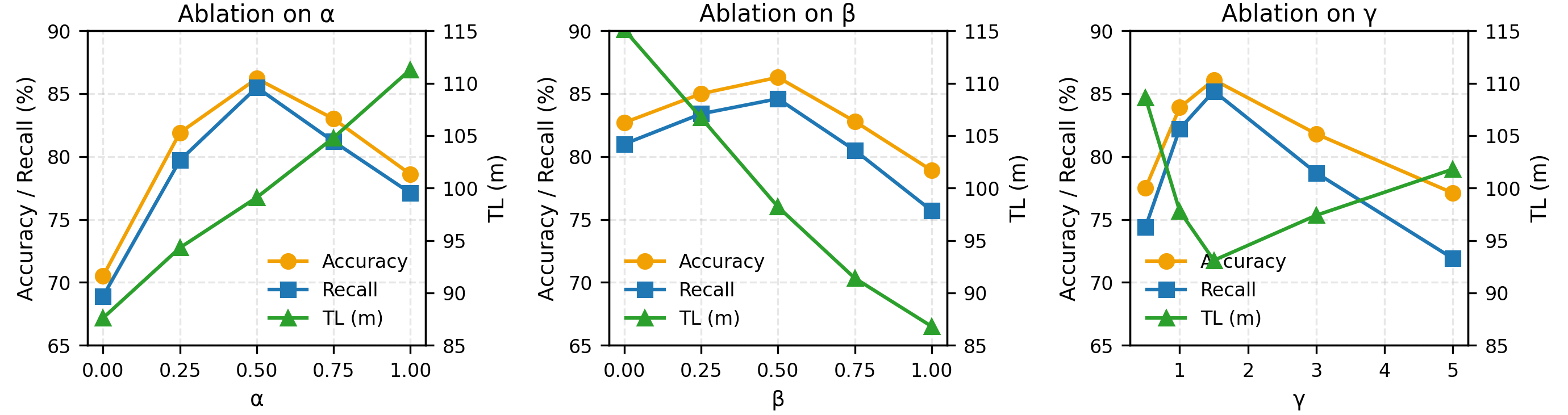}}
{\caption{
Ablation results on key parameters.
From left to right: $\alpha$, $\beta$, and $\gamma$.
Left axis: Accuracy / Recall (\%); Right axis: Trajectory Lengt (m).
}
\label{fig:param_ablation}
}
% \vspace{-6pt}
\end{figure}
As shown in Fig.~\ref{fig:param_ablation}, accuracy and recall follow bell-shaped trends, 
peaking near $\alpha{=}\beta{=}0.5$ and $\gamma{=}1.5$, 
which indicates balanced spatial–semantic fusion and stable LLM reasoning. 
The trajectory length slightly decreases with higher spatial weighting, 
showing that moderate spatial bias improves efficiency without sacrificing semantic precision.

\subsection{Real-world Environmental Experiments}

\begin{figure}[!htbp]
  \centering
  \includegraphics[scale=0.8]{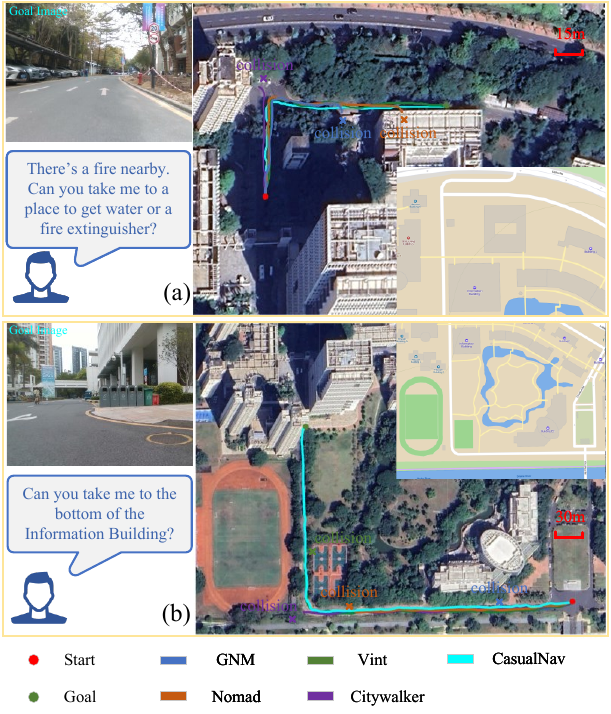}
  \caption{Experiments under different distance scales in real-world scenarios. }
  \label{figure:realexp}
\end{figure}

We deployed CausalNav on robotic platforms and conducted real-world comparisons with ViNT, NoMaD, GNM, and CityWalker, as shown in Fig.~\ref{figure:realexp}. Experiments were performed under two conditions. The first scenario (Fig.~\ref{figure:realexp}(a)) involved short-range navigation (130~m) with object-level instructions, where only ViNT and CausalNav succeeded. The second scenario (Fig.~\ref{figure:realexp}(b)) tested long-range navigation (512~m) using building-level instructions; only CausalNav completed the task, while others failed due to collisions.Simulation results are further validated in real-world settings. Both ViNT and CausalNav succeed in completing navigation tasks over distances of approximately 100 meters; however, only CausalNav is able to successfully perform long-range navigation in highly dynamic outdoor environments exceeding 500 meters. In particular, CityWalker exhibits significantly lower performance in real-world experiments compared to simulation. We observed that CityWalker is highly sensitive to lighting conditions and environmental changes, leading to unstable trajectory predictions and inadequate handling of dynamic objects. Although it demonstrates a high success rate in long-range tasks under simulation, the real world introduces additional complexities not captured in simulation. In particular, minor collisions that are tolerable in simulation may lead to task failure in real-world deployments. These findings confirm the robustness of CausalNav for long-distance semantic navigation in complex and dynamic urban environments.

\section{Conclusion and Future Work}

This work presents \textit{CausalNav}, the first scene-graph-based semantic navigation framework for dynamic outdoor environments. The system integrates LLM-constructed multi-level \textit{Embodied Graph} with retrieval-augmented reasoning and hierarchical planning, enabling long-range and open-vocabulary navigation with real-time adaptability. Extensive experiments in simulation and real-world settings demonstrate the effectiveness of our approach in improving outdoor navigation robustness and efficiency.

\textbf{Discussion and Limitations.} While effective in dynamic scenes, CausalNav still faces limitations in scalability, robustness under extreme lighting or weather, and long-horizon consistency. Compared with previous graph-memory systems such as \textit{EmbodiedRAG}~\cite{booker2024embodiedrag} and \textit{NavRAG}~\cite{wang2025navrag}, CausalNav extends the RAG paradigm to dynamic outdoor navigation via bidirectional LLM–RAG reasoning and real-time multi-level graph updates, achieving open-vocabulary understanding and continuous spatial-semantic adaptation.

{\textbf{Future Work.} We plan to enhance graph compression and memory recall mechanisms to improve scalability, explore multimodal fusion for better robustness under extreme conditions, and extend the system toward long-term autonomous exploration and lifelong learning.}

{
    % \balance
    \bibliographystyle{IEEEtran}
    \bibliography{IEEEabrv, reference/reference}
}

\end{document}